\documentclass[10pt]{article}

\pdfoutput=1

\usepackage[accepted]{rlj}

\usepackage{amssymb}
\usepackage{mathtools}
\usepackage{mathrsfs}
\usepackage{graphicx}
\usepackage{subcaption}
\usepackage{paracol}
\usepackage{url}
\usepackage{lipsum}

\usepackage[strict]{changepage}
\usepackage[space]{grffile}

\usepackage{algorithmic}
\usepackage{algorithm}
\usepackage{amsfonts}
\usepackage{amsmath}
\usepackage{amsthm}
\usepackage{booktabs}
\usepackage{caption}
\usepackage{hyperref}
\usepackage{microtype}
\usepackage{multicol}
\usepackage{multirow}
\usepackage{nicefrac}
\usepackage{relsize}
\usepackage{thm-restate}
\usepackage{thmtools}
\usepackage{tikz}
\usepackage{wrapfig}
\usepackage{xcolor}
\usepackage{xspace}
\usepackage{siunitx}

\definecolor{C0}{HTML}{1F77B4}
\definecolor{C1}{HTML}{FF7F0E}
\definecolor{C2}{HTML}{2CA02C}
\definecolor{C3}{HTML}{D62728}
\definecolor{C4}{HTML}{9467BD}
\definecolor{C5}{HTML}{8C564B}
\definecolor{C6}{HTML}{E377C2}
\definecolor{C7}{HTML}{7F7F7F}
\definecolor{C8}{HTML}{BCBD22}
\definecolor{C9}{HTML}{17BECF}

\newcommand{\uninformed}{\textcolor{black}{Uninformed Dreamer}\xspace}
\newcommand{\informed}{\textcolor{black}{Informed Dreamer}\xspace}
\newcommand{\scaffolder}{\textcolor{black}{Scaffolder}\xspace}
\newcommand{\reinformed}{\textcolor{black}{Reinformed Dreamer}\xspace}

\allowdisplaybreaks

\usetikzlibrary{arrows.meta, positioning}

\def\Eta{\mathrm H}

\DeclareMathOperator*{\argmax}{arg\,max}

\definecolor{DarkBlue}{HTML}{36587E}

\declaretheoremstyle[headpunct={\normalfont.}]{theoremstyle}

\makeatletter
\newcommand{\owntag}[2][\relax]{%
    \ifx#1\relax\relax\def\owntag@name{#2}\else\def\owntag@name{#1}\fi%
    \refstepcounter{equation}\tag{#2, \theequation}%
    \expandafter\ltx@label\expandafter{eq:\owntag@name}%
    \edef\@currentlabel{#2, \theequation}\expandafter\ltx@label\expandafter{Eq:\owntag@name}%
    \def\@currentlabel{#2}\expandafter\ltx@label\expandafter{tag:\owntag@name}%
}
\makeatother

\makeatletter
\newcommand*\wt[1]{\mathpalette\wthelper{#1}}
\newcommand*\wthelper[2]{%
    \hbox{\dimen@\accentfontxheight#1%
    \accentfontxheight#11.2\dimen@
    $\m@th#1\widetilde{#2}$%
    \accentfontxheight#1\dimen@
    }%
}
\newcommand*\accentfontxheight[1]{%
    \fontdimen5\ifx#1\displaystyle
    \textfont
    \else\ifx#1\textstyle
    \textfont
    \else\ifx#1\scriptstyle
    \scriptfont
    \else
    \scriptscriptfont
    \fi\fi\fi3
}
\makeatother

\def\titletext{Reinformed Dreamer: An Asymmetric World Model Efficiently Trained through Latent Guidance}
\title{\titletext}
\setrunningtitle{\titletext}

\author{%
    Gaspard Lambrechts\textsuperscript{1,2},
    Adrien Bolland\textsuperscript{3},
    Daniel Ebi\textsuperscript{4},
    Damien Ernst\textsuperscript{3}
}

\emails{
    gaspard.lambrechts@mcgill.ca
}

\affiliations{
    $^{1}$\textbf{McGill University, Canada} \\
    $^{2}$\textbf{Mila Québec Artificial Intelligence Institute, Canada} \\
    $^{3}$\textbf{Université de Liège, Belgium} \\
    $^{4}$\textbf{Karlsruhe Institute of Technology, Germany} \\
}


\begin{document}

\maketitle

\begin{abstract}
    Much like humans benefit from guidance while learning, reinforcement learning algorithms may benefit from additional supervision beyond rewards.
    Leveraging additional information during training to learn better representations and behaviors has been the focus of asymmetric reinforcement learning.
    This learning paradigm has proven effective under partial observability when additional state information is available, but also under full observability when more refined state information is available.
    Focusing on model-based reinforcement learning, we study the effect of asymmetric learning on observation representations and on privileged information representations.
    First, we identify a limitation in the privileged information representations learned by an asymmetric model-based algorithm known as the Informed Dreamer.
    Then, we propose a novel asymmetric representation learning objective using latent guidance, resulting in a new algorithm called the Reinformed Dreamer.
    Experiments across several benchmarks show a more consistent improvement over Dreamer than previous asymmetric approaches.
\end{abstract}

\section{Introduction} \label{sec:introduction}

In the field of artificial intelligence, intelligence is usually understood as the ability to make decisions, based on perception, in order to achieve goals \citep{mccarthy1998what, russell2010artificial}.
In other words, intelligence is about perceiving and abstracting past information about the world for then acting on its future execution, in the perspective of achieving objectives \citep{lambrechts2025reinforcement}.
Reinforcement learning (RL) is an appealing approach for learning intelligent behaviors, notably because it makes few assumptions about the decision problem \citep{sutton1998reinforcement}.
In its purest form, the promise of an RL algorithm is to learn an optimal behavior from interaction with an unknown decision process.
In general, the perception of the decision process consists of partial observations of its underlying state \citep{astrom1965optimal}, which defines a partially observable Markov decision process (POMDP).
Under partial observability, the optimal decision generally depends on the entire history of observations and past actions.
These histories grow unboundedly and demand an appropriate method to compress them into adequate representations.
Learning these abstractions, or representations, is known as representation learning, and has been one of the main focuses of RL in POMDP \citep{ni2024bridging}.

While learning from interaction and solely from partial observations of the decision process is a great promise, it is a restrictive learning paradigm.
As initially noticed in the supervised learning setting \citep{vapnik2009new}, additional information may be available during training to improve learning.
In RL, similar ideas have emerged under the name asymmetric RL \citep{pinto2017asymmetric}, motivated by the asymmetry of observability between training and execution.
The additional information may take the form of privileged observations from additional sensors, access to estimates of the state or to the simulator state, information about future states in hindsight, etc.
Recently, a line of work has focused on proposing theoretically grounded asymmetric learning objectives \citep{warrington2021robust, baisero2022unbiased, lambrechts2024informed, wang2023learning}.
Some of these methods, especially those leveraging an asymmetric critic, have demonstrated impressive empirical performance in various domains \citep{degrave2022magnetic, kaufmann2023champion, vasco2024super, hu2025real, durr2026outplaying}.
We provide an overview of asymmetric RL methods in \autoref{app:related}.

In the symmetric setting, model-based RL has proven an effective approach for solving decision-making problems, including under partial observability \citep{hafner2025mastering, samsami2024mastering}.
In model-based RL, a model of the decision process, known as a world model \citep{ha2018recurrent}, is learned from interaction samples.
In the Dreamer algorithm \citep{hafner2020dream, hafner2021mastering, hafner2025mastering}, which we will refer to as the \uninformed in the following, the world model is trained to recurrently predict a latent representation of the reward and next observation given the history of observations and actions.
Such world modeling objectives correspond to representation learning objectives that provide representations of the history that are sufficient for optimal control \citep{striebel1965sufficient}, also known as information states \citep{subramanian2022approximate}.
From the world model, and using an adequate policy, it is possible to generate synthetic trajectories of interaction using these latent representations, a process called imagination.
It has proved effective to use an actor-critic algorithm based on these synthetic trajectories in order to optimize the policy \citep{hafner2025mastering}.

Recently, model-based algorithms were adapted to the asymmetric setting to benefit from eventual additional information.
The \informed algorithm \citep{lambrechts2024informed} adapted the \uninformed \citep{hafner2025mastering} by introducing an asymmetric world model trained to predict a latent representation of the reward and next privileged information given the history of observations and actions, instead of a representation of the reward and next observation.
By ensuring that the privileged information embeds all relevant information from the observation, its representation is also a sufficient representation of the observation, and the imagination process remains valid.
This algorithm demonstrated an improved learning speed on various partially observable environments.
The \scaffolder algorithm \citep{hu2024privileged} outperformed the \informed by learning a privileged world model in addition.
The privileged world model learns a latent representation of the reward and next privileged information based on the history of privileged informations and past actions.
To obtain the representations based on the history of observations and past actions, needed during execution, an unprivileged world model is learned jointly.
During imagination, the privileged world model first generates latent representations based on the privileged history.
These representations are then decoded into observations and reencoded by the unprivileged world model.
This results in a costly imagination process, and one would ideally aim to achieve similar performance with a single world model and without explicitly decoding the observations.

This work makes several contributions towards this goal.
First, we show that the learning objective for the latent representation of the privileged information is problematic in the \informed.
Indeed, the asymmetric variational autoencoder uses a generally loose lower bound on the likelihood of the privileged information.
Second, we propose a proper learning objective for the latent representation of the privileged information by introducing a privileged variational encoder.
Since the privileged encoder cannot be used during execution, an unprivileged encoder is jointly learned through latent guidance.
This new world modeling objective defines the \reinformed algorithm, which features two encoders in a single world model, and whose imagination process remains unchanged.
Third, we show in several suites of environments that the \reinformed outperforms or matches previous approaches in terms of convergence speed and final performance.

\section{Informed Partially Observable Markov Decision Processes} \label{sec:background}

We model the asymmetry of observability between training and execution with the informed POMDP \citep{lambrechts2024informed}.
It is a generalization of the POMDP \citep{astrom1965optimal} that introduces the (privileged) information that is available during training.
Formally, an informed POMDP $\wt{\mathcal{P}}$ is a tuple $\wt{\mathcal{P}} = \allowbreak (\mathcal{S}, \mathcal{A}, \mathcal{I}, \mathcal{O}, P, T, R, \wt{I}, \wt{O}, \gamma)$, where $\mathcal{S}, \mathcal{A}, \mathcal{I}, \mathcal{O}$ are the state, action, information, and observation spaces, respectively.
$P \in \Delta(\mathcal{S})$ is the initial state distribution that gives the probability $P(s_0)$ of $s_0 \in \mathcal{S}$ being the initial state.
$T\colon \mathcal{S} \times \mathcal{A} \rightarrow \Delta(\mathcal{S})$ is the transition distribution that gives the probability $T(s_{t+1} | s_t, a_t)$ of $s_{t+1} \in \mathcal{S}$ being the state resulting from action $a_t \in \mathcal{A}$ in state $s_t \in \mathcal{S}$.
$R\colon \mathcal{S} \times \mathcal{A} \rightarrow \mathbb{R}$ is the reward function that gives the expected immediate reward $r_t = R(s_t, a_t)$ obtained when taking action $a_t \in \mathcal{A}$ in state $s_t \in \mathcal{S}$.
$\wt{I}$ is the information distribution that gives the probability $\wt{I}(i_t | s_t)$ of obtaining information $i_t \in \mathcal{I}$ in state $s_t \in \mathcal{S}$, and $\wt{O}$ is the observation distribution that gives the probability $\wt{O}(o_t | i_t)$ of obtaining observation $o_t \in \mathcal{O}$ given information $i_t \in \mathcal{I}$.
$\gamma \in [0, 1)$ is the discount factor that weighs the relative importance of future rewards.

\begin{wrapfigure}{r}{0.48\linewidth}
    \centering
    \vspace{-1.2em}
    \begin{tikzpicture}[font=\footnotesize, scale=1.0]
        \input{tikz/styles.tex}
        \draw[zone, opacity=0.30] (-0.4, -0.70) rectangle (4.4, -2.35) ;
\draw[zone, opacity=0.40] (-0.3, -1.75) rectangle (4.3, -2.25) ;

\input{tikz/trajectory}

\node[state] (s0) at (0, 0) {} ; \node[label] (s0t) at (s0) {$s$} ;
\node[state] (s1) at (2, 0) {} ; \node[label] (s1t) at (s1) {$s$} ;
\node[state] (s2) at (4, 0) {} ; \node[label] (s2t) at (s2) {$s$} ;

\draw[arrow] (s0) to (i0) ;
\draw[arrow] (i0) to (o0) ;

\draw[arrow] (s0) to (r0) ; \draw[arrow] (a0) to (r0) ;
\draw[arrow] (s0) to (s1) ; \draw[arrow] (a0) to (s1) ;

\draw[arrow] (s1) to (i1) ;
\draw[arrow] (i1) to (o1) ;

\draw[arrow] (s1) to (r1) ; \draw[arrow] (a1) to (r1) ;
\draw[arrow] (s1) to (s2) ; \draw[arrow] (a1) to (s2) ;

\draw[arrow] (s2) to (i2) ;
\draw[arrow] (i2) to (o2) ;
        \input{tikz/dynamics.tex}
        \input{tikz/legend.tex}
    \end{tikzpicture}
    \caption{Bayesian graph of an informed POMDP \citep{lambrechts2024informed}.}
    \vspace{-1em}
    \label{fig:ipomdp}
\end{wrapfigure}

Taking a sequence of $t$ actions in the informed POMDP conditions its execution and provides samples $(i_0, o_0, a_0, r_0, \dots, i_t, o_t)$ during training.
During execution, however, the privileged informations and rewards are not available anymore.
As a result, interacting with the informed POMDP becomes equivalent to interacting with its underlying execution POMDP, defined as $\mathcal{P} = (\mathcal{S}, \mathcal{A}, \mathcal{O}, P, T, R, O, \gamma)$ where ${O(o_t | s_t) = \mathop{\mathbb{E}}_{\wt{I}(i | s_t)}[\wt{O}(o_t | i)]}$
is the observation distribution.
Taking a sequence of $t$ actions in the execution POMDP conditions its execution and provides the history $h_t = (o_0, a_0, \dots, o_t) \in \mathcal{H}$, where $\mathcal{H}$ is the set of histories of arbitrary length.
Note that the information samples $i_0, \dots, i_t$ and reward samples $r_0, \dots, r_{t-1}$ are not included.
The training and execution are summarized in \autoref{fig:ipomdp}.

A history-dependent stochastic policy $\eta \in \Eta$ is a mapping from histories to probability distributions over the action space, where $\Eta = \left\{ \eta\colon \mathcal{H} \rightarrow \Delta(\mathcal{A})\right\}$ is the set of such mappings.
A policy $\eta^*$ is said to be optimal for an informed POMDP when it is optimal in the underlying execution POMDP, i.e., when it maximizes the expected return
$
    J(\eta) = \mathbb{E}^{\eta} \!
    \left[
        \sum_{t=0}^\infty \gamma^t r_t
    \right]\!.
    \label{eq:return}
$
The RL objective for an informed POMDP is thus to find an optimal policy $\eta^* \in \argmax_{\eta \in \Eta} J(\eta)$ from online or offline interaction, that is from samples $\left\{(i_0^n, o_0^n, a_0^n, r_0^n, \dots, i_{t_n}^n, o_{t_n}^n)\right\}_{n=0}^{N-1}$.

\section{Informed Dreamer and Information Gap} \label{sec:informed}

In \autoref{subsec:informed}, we recall the \informed \citep{lambrechts2024informed}, an asymmetric model-based RL algorithm for informed POMDPs, and the \uninformed \citep{hafner2025mastering} is obtained as a special case of this algorithm when there is no additional information (i.e., when $i_t = o_t$).
In \autoref{subsec:loose}, we show that the learning objective of the world model of the \informed is optimized through a generally loose lower bound.

\subsection{Informed Dreamer} \label{subsec:informed}

\paragraph{World Model.}

In an informed POMDP, an informed world model is a model $p_\phi(r_t, i_{t+1} | h_t, a_t)$ of the history-dependent distribution $p(r_t, i_{t+1} | h_t, a_t)$.
In the Informed Dreamer, the distribution $p_\phi(r_t, i_{t+1} | h_t, a_t)$ is implemented by a particular latent variable model (LVM), known as a recurrent state space model (RSSM).
More precisely, for an initial hidden state $z_0 = 0$, at any $t \geq 0$,
\begin{align}
    e_k &\sim q_\phi^e(e_k | z_k, o_k), \; k \leq t, \owntag[rssm_encoder]{encoder} \\
    z_{k+1} &= u_\phi(z_k, e_k, a_k), \; k \leq t, \owntag[rssm_recurrence]{recurrence} \\
    \hat{e}_{t+1} &\sim p_\phi^{\hat{e}}(\hat{e}_{t+1} | z_{t+1}), \owntag[rssm_prior]{prior} \\
    \hat{r}_{t} &\sim p_\phi^r(\hat{r}_{t} | z_{t+1}, \hat{e}_{t+1}), \owntag[rssm_reward]{reward decoder} \\
    \hat{i}_{t+1} &\sim p_\phi^i(\hat{i}_{t+1} | z_{t+1}, \hat{e}_{t+1}). \owntag[rssm_information]{information decoder}
\end{align}
This LVM uses latent variables $(e_{0:t}, \hat e_{t+1})$ and its prior $p_\phi(e_{0:t}, \hat e_{t+1} | h_t, a_t)$ is given by:
\begin{align}
    q_\phi(e_{0:t} | h_t) &= \textstyle \prod_{k=0}^{t} q_\phi^e(e_k | z_k, o_k), \label{eq:pre_variational_posterior} \\
    p_\phi(e_{0:t}, \hat e_{t+1} | h_t, a_t) &= q_\phi(e_{0:t} | h_t) p_\phi^{\hat{e}}(\hat e_{t+1} | z_{t+1}) \label{eq:prior},
\end{align}
where $z_{k+1} = u_\phi(z_{k}, e_{k}, a_{k})$ for $k \leq t$.
The distribution $p_\phi(r_t, i_{t+1} | h_t, a_t)$ is thus modeled by:
\begin{align}
    p_\phi(r_t, i_{t+1} | h_t, a_t) &= \mathop{\mathbb{E}}_{p_\phi(e_{0:t}, \hat e_{t+1} | h_t, a_t)} [p_\phi(r_t, i_{t+1} | z_{t+1}, \hat{e}_{t+1})] \\
    &= \mathop{\mathbb{E}}_{p_\phi(e_{0:t}, \hat e_{t+1} | h_t, a_t)} [ p_\phi^r(r_t | z_{t+1}, \hat{e}_{t+1}) p_\phi^i(i_{t+1} | z_{t+1}, \hat{e}_{t+1}) ], \label{eq:lvm}
\end{align}
where $z_{k+1} = u_\phi(z_{k}, e_{k}, a_{k})$ for $k \leq t$.

\begin{figure}[ht]
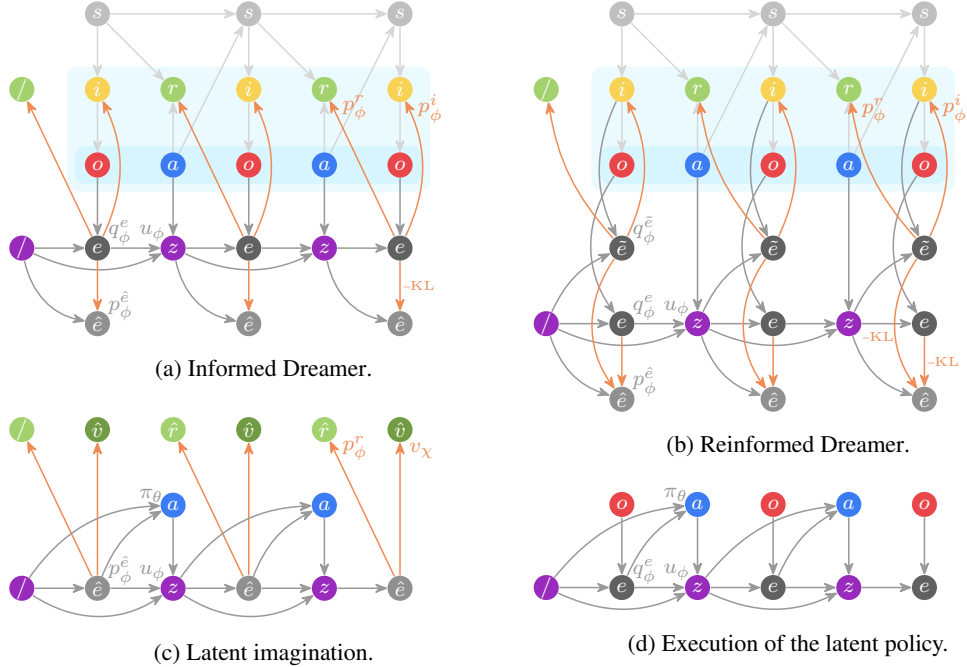

    \centering
    \begin{subfigure}[T]{0.49\linewidth}
        \begin{tikzpicture}[font=\footnotesize, scale=1.0]
            \input{tikz/styles.tex}
            \begin{scope}[transparency group, opacity=0.4]
                \draw[zone, opacity=0.30] (-0.4, -0.70) rectangle (4.4, -2.35) ;
\draw[zone, opacity=0.40] (-0.3, -1.75) rectangle (4.3, -2.25) ;

\input{tikz/trajectory}

\node[state] (s0) at (0, 0) {} ; \node[label] (s0t) at (s0) {$s$} ;
\node[state] (s1) at (2, 0) {} ; \node[label] (s1t) at (s1) {$s$} ;
\node[state] (s2) at (4, 0) {} ; \node[label] (s2t) at (s2) {$s$} ;

\draw[arrow] (s0) to (i0) ;
\draw[arrow] (i0) to (o0) ;

\draw[arrow] (s0) to (r0) ; \draw[arrow] (a0) to (r0) ;
\draw[arrow] (s0) to (s1) ; \draw[arrow] (a0) to (s1) ;

\draw[arrow] (s1) to (i1) ;
\draw[arrow] (i1) to (o1) ;

\draw[arrow] (s1) to (r1) ; \draw[arrow] (a1) to (r1) ;
\draw[arrow] (s1) to (s2) ; \draw[arrow] (a1) to (s2) ;

\draw[arrow] (s2) to (i2) ;
\draw[arrow] (i2) to (o2) ;
            \end{scope}
            \def\rec{-3.1}
            \input{tikz/initial.tex}
            \input{tikz/recurrence.tex}
            \def\bend{33}
            \def\pos{-4.1}
            \def\below{25}
            \input{tikz/prior.tex}
            \input{tikz/encoder.tex}
            \input{tikz/trajectory.tex}
            \node[reward] (r-1) at (-1, -1) {} ; \node[label] (r-1t) at (r-1) {$/$} ;

\draw[loss] (e0) to (r-1) ;
\draw[loss] (e1) to (r0) ;
\draw[loss] (e2) to node[pos=0.95, right, align=left] {$p_\phi^r$} (r1) ;

\draw[loss] (e0) to[bend right = 25] (i0) ;
\draw[loss] (e1) to[bend right = 25] (i1) ;
\draw[loss] (e2) to[bend right = 25] node[pos=0.95, right, align=left] {$p_\phi^i$} (i2) ;

\draw[loss] (e0) to (p0) ;
\draw[loss] (e1) to (p1) ;
\draw[loss] (e2) to node[pos=0.50, right=-0.08cm, align=left] {\tiny $\text{--}\!\operatorname{KL}$} (p2) ;
        \end{tikzpicture}
        \caption{\informed.}
        \label{fig:informed}
    \end{subfigure}
    \begin{subfigure}[T]{0.49\linewidth}
        \begin{tikzpicture}[font=\footnotesize, scale=1.0]
            \input{tikz/styles.tex}
            \begin{scope}[transparency group, opacity=0.4]
                \draw[zone, opacity=0.30] (-0.4, -0.70) rectangle (4.4, -2.35) ;
\draw[zone, opacity=0.40] (-0.3, -1.75) rectangle (4.3, -2.25) ;

\input{tikz/trajectory}

\node[state] (s0) at (0, 0) {} ; \node[label] (s0t) at (s0) {$s$} ;
\node[state] (s1) at (2, 0) {} ; \node[label] (s1t) at (s1) {$s$} ;
\node[state] (s2) at (4, 0) {} ; \node[label] (s2t) at (s2) {$s$} ;

\draw[arrow] (s0) to (i0) ;
\draw[arrow] (i0) to (o0) ;

\draw[arrow] (s0) to (r0) ; \draw[arrow] (a0) to (r0) ;
\draw[arrow] (s0) to (s1) ; \draw[arrow] (a0) to (s1) ;

\draw[arrow] (s1) to (i1) ;
\draw[arrow] (i1) to (o1) ;

\draw[arrow] (s1) to (r1) ; \draw[arrow] (a1) to (r1) ;
\draw[arrow] (s1) to (s2) ; \draw[arrow] (a1) to (s2) ;

\draw[arrow] (s2) to (i2) ;
\draw[arrow] (i2) to (o2) ;
            \end{scope}
            \def\rec{-4.1}
            \input{tikz/initial.tex}
            \input{tikz/recurrence.tex}
            \def\pos{-5.1}
            \def\bend{27}
            \def\below{25}
            \input{tikz/prior.tex}
            \input{tikz/unprivileged.tex}
            \node[encoded] (e0)   at (0.0, -3.1) {} ; \node[label] (e0t) at (e0) {$\tilde e$} ;
\node[encoded] (e1)    at (2.0, -3.1) {} ; \node[label] (e1t) at (e1) {$\tilde e$} ;
\node[encoded] (e2)    at (4.0, -3.1) {} ; \node[label] (e2t) at (e2) {$\tilde e$} ;

\draw[arrow] (z0) to[bend left = 20] (e0) ; % \draw[arrow] (a-1) to (e-1) ;
% \draw[arrow] (o0) to (e0) ;
\draw[arrow] (i0) to[bend right = 25] (e0) ;
\draw[arrow] (z1) to[bend left = 20] (e1) ; % \draw[arrow] (a0) to (e0) ;
% \draw[arrow] (o1) to (e1) ;
\draw[arrow] (i1) to[bend right = 25] (e1) ;
\draw[arrow] (z2) to[bend left = 20] (e2) ; % \draw[arrow] (a1) to (e1) ;
% \draw[arrow] (o2) to (e2) ;
\draw[arrow] (i2) to[bend right = 25] (e2) ;

% \draw[arrow] (a-1) to (z0) ;
\draw[arrow] (z0) to[bend right=25] (z1) ; \draw[arrow] (u0) to (z1) ; \draw[arrow] (a0) to (z1) ;
\draw[arrow] (z1) to[bend right=25] (z2) ; \draw[arrow] (u1) to (z2) ; \draw[arrow] (a1) to (z2) ;
% \draw[arrow] (z1) to[bend right=25] (z2) ; \draw[arrow] (e1) to (z2) ;

\node[distribution, above right=-0.2cm and -0.1cm of e0] {$q_\phi^{\tilde e}$} ;
            \input{tikz/trajectory.tex}
            \node[reward] (r-1) at (-1, -1) {} ; \node[label] (r-1t) at (r-1) {$/$} ;

\draw[loss] (e0) to[bend left=16] (r-1) ;
\draw[loss] (e1) to[bend left=16] (r0) ;
\draw[loss] (e2) to[bend left=16] node[pos=0.95, right, align=left] {$p_\phi^r$} (r1) ;

\draw[loss] (e0) to[bend right = 25] (i0) ;
\draw[loss] (e1) to[bend right = 25] (i1) ;
\draw[loss] (e2) to[bend right = 25] node[pos=0.95, right, align=left] {$p_\phi^i$} (i2) ;

\draw[loss] (e0) to[bend right=35] (p0) ;
\draw[loss] (e1) to[bend right=35] (p1) ;
\draw[loss] (e2) to[bend right=35] node[pos=0.60, left=-0.1cm, align=right] {\tiny $\text{--}\!\operatorname{KL}$} (p2) ;

\draw[loss] (u0) to (p0) ;
\draw[loss] (u1) to (p1) ;
\draw[loss] (u2) to node[pos=0.50, right=-0.08cm, align=left] {\tiny $\text{--}\!\operatorname{KL}$} (p2) ;
        \end{tikzpicture}
        \caption{\reinformed.}
        \label{fig:reinformed}
    \end{subfigure}
    \begin{subfigure}[T]{0.49\linewidth}
        \vspace{-2em}
        \begin{tikzpicture}[font=\footnotesize, scale=1.0]
            \input{tikz/styles.tex}
            \def\rec{-3.1}
            \input{tikz/initial.tex}
            \input{tikz/recurrence.tex}
            \def\bend{0}
            \def\pos{-3.1}
            \def\below{25}
            \input{tikz/prior.tex}
            \node (e0) at (p0) {};
            \node (e1) at (p1) {};
            \node (e2) at (p2) {};
            \input{tikz/imagined.tex}
            \def\bendone{26}
            \def\bendtwo{21}
            \draw[arrow, bend left = \bendone] (z0) to (a0) ;
\draw[arrow, bend left = \bendtwo] (e0) to (a0) ;
\draw[arrow, bend left = \bendone] (z1) to (a1) ;
\draw[arrow, bend left = \bendtwo] (e1) to (a1) ;

\node[distribution, above left=-0.20 and -0.0 of a0, align=right] {$\pi_\theta$} ;
        \end{tikzpicture}
        \caption{Latent imagination.}
        \label{fig:imagination}
    \end{subfigure}
    \begin{subfigure}[T]{0.49\linewidth}
        \vspace{0.6em}
        \begin{tikzpicture}[font=\footnotesize, scale=1.0]
            \input{tikz/styles.tex}
            \def\rec{-3.1}
            \input{tikz/initial.tex}
            \input{tikz/recurrence.tex}
            \input{tikz/history.tex}
            \input{tikz/encoder.tex}
            \def\bendone{22}
            \def\bendtwo{18}
            \draw[arrow, bend left = \bendone] (z0) to (a0) ;
\draw[arrow, bend left = \bendtwo] (e0) to (a0) ;
\draw[arrow, bend left = \bendone] (z1) to (a1) ;
\draw[arrow, bend left = \bendtwo] (e1) to (a1) ;

\node[distribution, above left=-0.20 and -0.0 of a0, align=right] {$\pi_\theta$} ;
        \end{tikzpicture}
        \caption{Execution of the latent policy.}
        \label{fig:execution}
    \end{subfigure}
    \caption{Training, imagination and execution of the \informed and \reinformed. In all figures, the dependence of the decoders and latent critic on the last hidden state is omitted.}
    \label{fig:illustrations}
\end{figure}

To learn the parameters $\phi$ of this model $p_\phi(r_t, i_{t+1} | h_t, a_t)$, we may want to minimize the KL divergence from the true distribution, which is known to be equivalent to maximizing the log-likelihood of samples from the true distribution.
Unfortunately, the log-likelihood $\log p_\phi(r_t, i_{t+1} | h_t, a_t)$ of such a model is not easy to estimate.
Inspired by the evidence lower bound (ELBO) in variational inference, the Informed Dreamer proposes instead to maximize the informed ELBO,
\begin{align}
    &\wt{\operatorname{ELBO}}_\phi(r_t, i_{t+1} | h_t, a_t)
    \nonumber \\
    &\qquad
    = \mathop{\mathbb{E}}_{\wt{O}(o_{t+1} | i_{t+1})} \bigg[
        \mathop{\mathbb{E}}_{q_\phi(e_{0:t}, e_{t+1} | h_t, a_t, o_{t+1})} \bigg[
                \log p_\phi^r(r_t | z_{t+1}, e_{t+1})
                +
                \log p_\phi^i(i_{t+1} | z_{t+1}, e_{t+1})
            \nonumber \\
            &\qquad\qquad\qquad\qquad\qquad\qquad\qquad
            - \operatorname{KL}_{e_{t+1}'}(q_\phi^{e}(e_{t+1}' | z_{t+1}, o_{t+1}) \parallel p_\phi^{\hat{e}}(e_{t+1}' | z_{t+1}))
        \bigg]
    \bigg],
    \label{eq:informed_elbo}
\end{align}
where $z_{k+1} = u_\phi(z_k, e_k, a_k)$ for $k \leq t$ and $q_\phi(e_{0:t}, e_{t+1} | h_t, a_t, o_{t+1}) = q_\phi(e_{0:t+1} | h_{t+1})$ is the variational distribution, following equation~\eqref{eq:pre_variational_posterior}.
As proven in \autoref{app:informed_elbo}, the informed ELBO is a valid lower bound on the log-likelihood: $\smash{\wt{\operatorname{ELBO}}_\phi(r_t, i_{t+1} | h_t, a_t) \leq \log p_\phi(r_t, i_{t+1} | h_t, a_t)}$.
The resulting world modeling objective is:
\begin{align}
    \max_{\phi} \mathop{\mathbb{E}}_{p(h_t, a_t, r_t, i_{t+1})} [ \wt{\operatorname{ELBO}}_\phi(r_t, i_{t+1} | h_t, a_t) ].
\end{align}
The world model architecture and its learning objective are illustrated in \autoref{fig:informed}.
Note that it also uses symlog predictions, a discrete latent space with straight-through gradients through categorical samples, and KL balancing with free bits as detailed in \autoref{app:details}.

\paragraph{Latent Policy.}

By selecting a latent policy $\pi_\theta(a_t | z_t, e_t)$, it is possible to generate sequences of interactions without reconstructing the observation nor the information.
Indeed, starting from the initial state $z_0 = 0$, we sample the initial latent variable $e_0 \sim p_\phi^{\hat e}(e_0 | z_0)$.
Then, for any $t \geq 0$, we select the action $a_t \sim \pi_\theta(a_t | z_t, e_t)$, update the hidden state to $z_{t+1} = u_\phi(z_t, e_t, a_t)$, and sample the next latent $e_{t+1} \sim p_\phi^{\hat e}(e_{t+1} | z_{t+1})$ from which we obtain the immediate reward $r_t \sim p_\phi^r(r_t | z_{t+1}, e_{t+1})$.
In practice, the policy is optimized to maximize the imagined return by an on-policy actor-critic algorithm with a latent critic $v_\chi(z_t, e_t)$, using policy gradient ascent based on the latent trajectories.
The latent imagination procedure is illustrated in \autoref{fig:imagination}.

While the latent variables are available during imagination, only the observations are available at execution.
As a result, the latent policy $\pi_\theta(a_t | z_t, e_t)$ cannot be deployed directly.
Instead, we infer the latent variables using the encoder $q_\phi(e_{0:t} | h_t) = \prod_{k=0}^t q_\phi^e(e_k | z_k, o_k)$ where $z_{k+1} = u_\phi(z_k, e_k, a_k)$ for $k < t$, and deploy the history-dependent policy $\eta_{\phi, \theta}(a_t | h_t)$, given by:
\begin{align}
    \eta_{\phi, \theta}(a_t | h_t) = \mathop\mathbb{E}_{q_\phi(e_{0:t} | h_t)} [ \pi_\theta(a_t | z_t, e_t) ].
\end{align}
The policy execution is illustrated in \autoref{fig:execution}.

\subsection{Looseness of the Informed Evidence Lower Bound} \label{subsec:loose}

When deriving the informed ELBO in equation~\eqref{eq:informed_elbo} based on an observational encoder $q_\phi(e_k | z_k, o_k)$, we obtain a loose lower bound on the log-likelihood $\log p_\phi(r_t, i_{t+1} | h_t, a_t)$. Indeed, we have:
\begin{align}
    &\wt{\operatorname{ELBO}}_{\phi}(r_t, i_{t+1} | h_t, a_t)
    =
    \log p_\phi(r_t, i_{t+1} | h_t, a_t)
    \nonumber \\
    &\qquad
    -
    \mathop{\mathbb{E}}_{\wt{O}(o_{t+1} | i_{t+1})} \big[
        \operatorname{KL}_{e_{0:t+1}}(q_\phi(e_{0:t+1} | h_t, a_t, o_{t+1}) \parallel p_\phi(e_{0:t+1} | h_t, a_t, r_t, i_{t+1}))
    \big],
    \label{eq:untightness}
\end{align}
where $q_\phi(e_{0:t+1} | h_t, a_t, o_{t+1})$ is the variational distribution and $p_\phi(e_{0:t+1} | h_t, a_t, r_t, i_{t+1})$ is the true posterior of the LVM defined by equation~\eqref{eq:lvm}.
The proof is given in \autoref{app:informed_gap}.

The KL term cannot be made arbitrarily small in general.
First, the recurrent factorization of the variational distribution $q_\phi(e_{0:t+1} | h_t, a_t, o_{t+1})$ using the encoder $q_\phi^e(e_k | z_k, o_k)$ assumes that $e_k$ is conditionally independent of the future given $h_k$, for all $k \leq t$.
It is not true in general that the true posterior $p_\phi(e_{0:t+1} | h_t, a_t, r_t, i_{t+1})$ of the LVM has such conditional independence.
This is known as the conditioning gap, which is inherent to the RSSM architecture \citep{becker2022uncertainty}.

In addition, the variational distribution $q_\phi(e_{0:t+1} | h_t, a_t, o_{t+1})$ is conditioned on less information than the true posterior $p_\phi(e_{0:t+1} | h_t, a_t, r_t, i_{t+1})$, as it omits the reward $r_t$ and the next information $i_{t+1}$.
It is not true in general that the true posterior $p_\phi(e_{0:t+1} | h_t, a_t, r_t, i_{t+1})$ is independent of the reward $r_t$ and next information $i_{t+1}$ given the observation $o_{t+1}$, along with the history $h_t$ and action $a_t$.
We call this the information gap.
As a consequence, the informed world model implemented by the LVM at equation~\eqref{eq:lvm} may be optimized through a loose lower bound, which can hinder learning.

In the particular case where the observation $o_{t+1}$ does not contain any information about the information $i_{t+1}$ (and the reward $r_t$), the ELBO will necessarily remain loose, and the distribution will be poorly approximated.
Without any information about the information $i_{t+1}$ (and the reward $r_t$) coming from the encoder, the learned decoder distribution will result in the best decoder fit, a problem akin to posterior collapse \citep{he2018lagging}.
In practice, the decoder class is a Gaussian of fixed variance, such that the best fit only encodes the conditional mean.
Note that it is not a problem for the reward distribution, as we target the average return.

In the particular case where the observation $o_{t+1}$ contains all information about the information $i_{t+1}$ (and the reward $r_t$), the ELBO can become tight, up to the conditional gap of \citet{becker2022uncertainty}, and the distribution may be correctly approximated.
Since the informed ELBO can only be tight when the observation $o_{t+1}$ contains all information about the information $i_{t+1}$ (and the reward $r_t$), we conclude that the \informed does not benefit from getting more information about the state $s_{t+1}$ in the information $i_{t+1}$ than in the observation $o_{t+1}$.
Instead, we hypothesize that it may benefit from having less irrelevant information in the information $i_{t+1}$ than in the observation $o_{t+1}$.

\section{Reinformed Dreamer} \label{sec:reinformed}

To reduce the tightness gap of equation~\eqref{eq:untightness}, we introduce the \reinformed that has a privileged encoder to provide a tighter ELBO, and an unprivileged encoder trained by a latent guidance mechanism.
The privileged encoder $q_\phi^{\tilde{e}}(e_{t+1} | z_{t+1}, i_{t+1})$ is conditioned on the next information to reduce the information gap.
Note that it is not conditioned on the reward, as learning the conditional mean of the reward is sufficient for optimal control.
To deploy the policy, we also keep the encoder $q_\phi^e(e_k | z_k, o_k)$ for $k \leq t$, which we call the unprivileged encoder.
The resulting reinformed RSSM is:
\begin{align}
            e_k &\sim q_\phi^e(e_k | z_k, o_k), \; k \leq t, \owntag[irssm_unprivileged]{unprivileged encoder} \\
            z_{k+1} &= u_\phi(z_k, e_k, a_k), \; k \leq t, \owntag[irssm_recurrence]{recurrence} \\
            \tilde e_{t+1} &\sim q_\phi^{\tilde{e}}(\tilde e_{t+1} | z_{t+1}, i_{t+1}), \owntag[irssm_privileged]{privileged encoder} \\
            \hat{e}_{t+1} &\sim p_\phi^{\hat{e}}(\hat{e}_{t+1} | z_{t+1}), \owntag[irssm_prior]{prior} \\
            \hat{r}_{t} &\sim p_\phi^r(\hat{r}_{t} | z_{t+1}, \hat{e}_{t+1}), \owntag[irssm_reward]{reward decoder} \\
            \hat{i}_{t+1} &\sim p_\phi^i(\hat{i}_{t+1} | z_{t+1}, \hat{e}_{t+1}). \owntag[irssm_information]{information decoder}
\end{align}
This RSSM implements the same LVM as that of the Informed Dreamer at equation~\eqref{eq:lvm}, where the privileged encoder is not used.
However, to train this LVM, we propose to use the privileged encoder to tighten the ELBO, resulting in the reinformed ELBO:
\begin{align}
    &\operatorname{ELBO}_\phi(r_t, i_{t+1} | h_t, a_t)
    =
        \mathop{\mathbb{E}}_{q_\phi(e_{0:t}, \tilde e_{t+1} | h_t, a_t, i_{t+1})} \bigg[
                \log p_\phi^r(r_t | z_{t+1}, \tilde e_{t+1})
                +
                \log p_\phi^i(i_{t+1} | z_{t+1}, \tilde e_{t+1})
            \nonumber \\
            &\qquad\qquad\qquad\qquad\qquad\qquad\qquad\qquad
            - \operatorname{KL}_{\tilde e_{t+1}'}(q_\phi^{\tilde e}(\tilde e_{t+1}' | z_{t+1}, i_{t+1}) \parallel p_\phi^{\hat{e}}(\tilde e_{t+1}' | z_{t+1}))
        \bigg],
    \label{eq:reinformed_elbo}
\end{align}
where $q_\phi(e_{0:t}, \tilde e_{t+1} | h_t, a_t, i_{t+1}) = q_\phi(e_{0:t} | h_{t}) q_\phi^{\tilde{e}}(\tilde e_{t+1} | z_{t+1}, i_{t+1})$ is the variational distribution, with $q_\phi(e_{0:t} | h_t)$ given by equation~\eqref{eq:pre_variational_posterior}, and $z_{k+1} = u_\phi(z_k, e_k, a_k)$ for $k \leq t$.
The proof is given in \autoref{app:reinformed_elbo}.
Interestingly, this lower bound is tight when
$
    \operatorname{KL}_{e_{0:t}, \tilde e_{t+1}}(
        q_\phi(e_{0:t}, \tilde e_{t+1} | h_t, a_t, i_{t+1})
        \parallel
        p_\phi(e_{0:t}, \tilde e_{t+1} | h_t, a_t, r_t, i_{t+1})
    ) = 0,
$
where $p_\phi(e_{0:t}, \tilde e_{t+1} | h_t, a_t, r_t, i_{t+1})$ is the true posterior of the LVM defined by equation~\eqref{eq:lvm}.
The proof is in \autoref{app:reinformed_gap}.
More precisely, because the privileged variational distribution is conditioned on more information than the unprivileged variational distribution of the Informed Dreamer, the corresponding gap will be smaller than the tightness gap identified in equation~\eqref{eq:untightness}.
There may however remain a conditioning gap resulting from the recurrence of the variational distribution, as identified by \citet{becker2022uncertainty}.
To also train the unprivileged encoder~\eqref{eq:irssm_unprivileged} needed at execution as in the \informed, we use the following learning objective:
\begin{align}
    &\operatorname{ELBO}^+_{\phi}(r_t, i_{t+1} | h_t, a_t)
    =
        \mathop{\mathbb{E}}_{q_\phi(e_{0:t}, \tilde e_{t+1} | h_t, a_t, i_{t+1})} \Bigg[
                \log p_\phi^r(r_t | z_{t+1}, \tilde e_{t+1})
                +
                \log p_\phi^i(i_{t+1} | z_{t+1}, \tilde e_{t+1})
            \nonumber \\
    &\qquad\qquad\qquad\qquad\qquad\qquad
            - \operatorname{KL}_{\tilde e_{t+1}'}(q_\phi^{\tilde e}(\tilde e_{t+1}' | z_{t+1}, i_{t+1}) \parallel p_\phi^{\hat{e}}(\tilde e_{t+1}' | z_{t+1}))
    \nonumber \\
    &\qquad\qquad\qquad\qquad\qquad\qquad
    - \mathop{\mathbb{E}}_{\wt{O}(o_{t+1}|i_{t+1})} \Big[
        \operatorname{KL}_{e_{t+1}'}(q_\phi^{e}(e_{t+1}' | z_{t+1}, o_{t+1}) \parallel p_\phi^{\hat{e}}(e_{t+1}' | z_{t+1}))
    \Big]
        \Bigg].
    \label{eq:augmented_elbo}
\end{align}
As seen in equation~\eqref{eq:augmented_elbo}, the unprivileged encoder $q^e_\phi(e_t | z_t, o_t)$ is trained to match the prior $p_\phi^{\hat e}(e_t | z_t)$, and vice versa.
The prior $p_\phi^{\hat e}(e_t | z_t)$ itself is trained to match the privileged encoder $q^{\tilde e}_\phi(e_t | z_t, i_t)$, and vice versa.
We refer to this overall learning process for the prior, unprivileged encoder and privileged encoder as the latent guidance mechanism.

Finally, although not necessary, we propose to further increase the supervision signal on the unprivileged encoder by adding a reconstruction loss using the same decoders as for the privileged encoder.
The final world modeling objective for the Reinformed Dreamer algorithm is:
\begin{align}
    &\operatorname{ELBO}^{++}_{\phi}(r_t, i_{t+1} | h_t, a_t)
    =
        \mathop{\mathbb{E}}_{q_\phi(e_{0:t}, \tilde e_{t+1} | h_t, a_t, i_{t+1})} \Bigg[
                \log p_\phi^r(r_t | z_{t+1}, \tilde e_{t+1})
                +
                \log p_\phi^i(i_{t+1} | z_{t+1}, \tilde e_{t+1})
            \nonumber \\
    &\qquad\qquad\qquad\qquad\qquad\qquad\quad
            - \operatorname{KL}_{\tilde e_{t+1}'}(q_\phi^{\tilde e}(\tilde e_{t+1}' | z_{t+1}, i_{t+1}) \parallel p_\phi^{\hat{e}}(\tilde e_{t+1}' | z_{t+1}))
    \nonumber \\
    &\qquad\qquad\qquad
    + \mathop{\mathbb{E}}_{\wt{O}(o_{t+1}|i_{t+1})} \bigg[
        \mathop{\mathbb{E}}_{q_\phi^e(e_{t+1} | z_{t+1}, o_{t+1})} \Big[
                \log p_\phi^r(r_t | z_{t+1}, e_{t+1})
                +
                \log p_\phi^i(i_{t+1} | z_{t+1}, e_{t+1})
            \Big]
        \nonumber \\
    &\qquad\qquad\qquad\qquad\qquad\qquad\quad
        - \operatorname{KL}_{e_{t+1}'}(q_\phi^{e}(e_{t+1}' | z_{t+1}, o_{t+1}) \parallel p_\phi^{\hat{e}}(e_{t+1}' | z_{t+1}))
    \bigg]
        \Bigg].
    \label{eq:double_augmented_elbo}
\end{align}

The architecture and learning objectives of the \reinformed are illustrated in \autoref{fig:reinformed}.
As the \informed, it uses symlog predictions, a discrete latent space with the Gumbel softmax reparametrization trick, and KL balancing with free bits as detailed in \autoref{app:details}.
The imagination and execution procedures are unchanged and are illustrated in \autoref{fig:imagination} and \autoref{fig:execution}, respectively.

\section{Experiments} \label{sec:experiments}

We evaluate the \reinformed on three suites of informed POMDPs, where privileged information is available during training.
First, the Varying Mountain Hike environments \citep{igl2018deep, lambrechts2024informed} are navigation tasks where the observation is a noisy measurement of the altitude, while the privileged information is the exact position and orientation of the agent.
Second, the Pop Gym suite \citep{morad2023popgym} is a suite of partially observable environments in which the privileged information is the full state.
Finally, the Velocity Control suite \citep{tassa2018deepmind} is a version of the DeepMind Control Suite where only the velocity of the joints is observable, and the additional information is the exact position of the joints.
We compare against the \uninformed, the \informed, and the \scaffolder algorithm, using the same hyperparameters as in the original implementations, which are all based on the \href{https://github.com/danijar/dreamerv3}{DreamerV3} codebase.
All results are averaged over five seeds.
Other implementation details are reported in \autoref{app:details}.

\begin{figure}[ht]
    \centering
    \begin{subfigure}{\linewidth}
        \centering
        \includegraphics[width=0.68\linewidth]{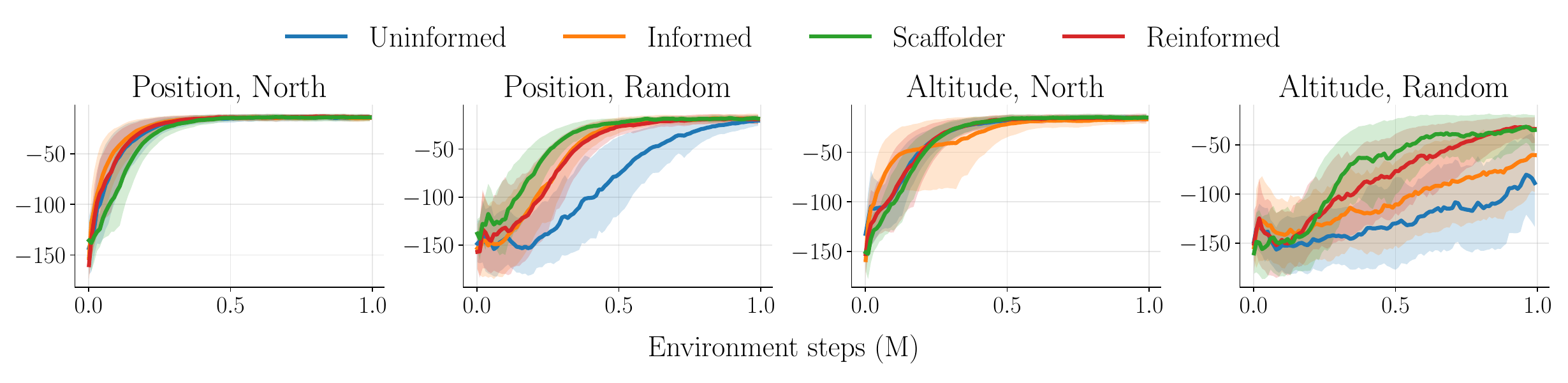}
        \caption{Varying Mountain Hike Environments}
        \vspace{1em}
        \label{fig:vmh}
    \end{subfigure}
    \begin{subfigure}{\linewidth}
        \centering
        \includegraphics[width=0.84\linewidth]{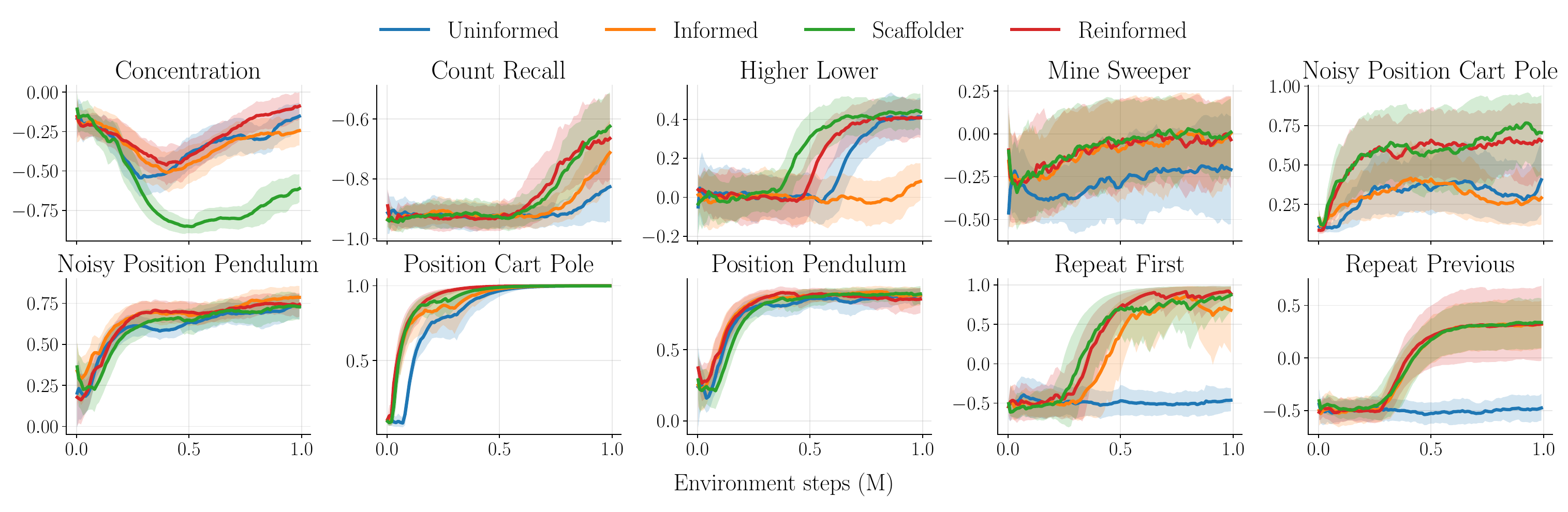}
        \caption{Pop Gym Environments}
        \vspace{1em}
        \label{fig:pop}
    \end{subfigure}
    \begin{subfigure}{\linewidth}
        \centering
        \includegraphics[width=\linewidth]{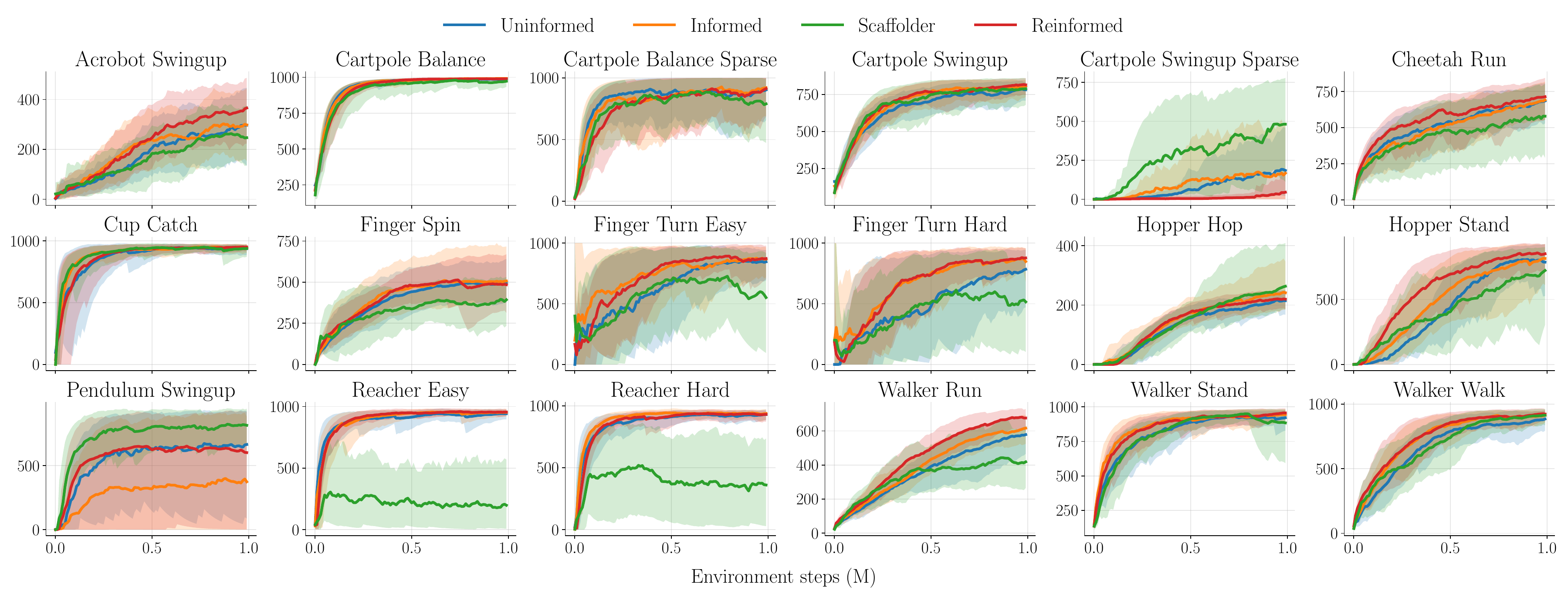}
        \caption{Velocity Control environments}
        \vspace{1em}
        \label{fig:velocity}
    \end{subfigure}
    \caption{Evolution of the cumulative reward over one million training steps across all environments. For each method, we report the minimum, mean and maximum performance over five training seeds.}
    \label{fig:results}
\end{figure}

As shown in \autoref{fig:results}, the \reinformed outperforms or matches the performance of the \uninformed and \informed in all but four environments.
The \reinformed also outperforms or matches the performance of the \scaffolder in most environments, except two Velocity Control environments, and two Varying Mountain Hike environments.
We consider these to be promising results that motivate further work on more challenging tasks.
As shown in \autoref{tab:runtimes}, the \reinformed is about 20\% slower to train than the \uninformed and \informed, versus approximately 80\% to 120\% for the \scaffolder and its two world models.

\begin{table}[ht]
\centering
\caption{Average runtime for training over one million steps across each suite for each algorithm.}
\label{tab:runtimes}
\begin{tabular}{l|cc|cc|cc}
\toprule
Algorithm & \multicolumn{2}{c|}{Varying Mountain Hike} & \multicolumn{2}{c|}{Pop Gym} & \multicolumn{2}{c}{Velocity Control} \\
\midrule
\uninformed      & \qty{2570.3}{\second} & 100.0 \% & \qty{2274.7}{\second} & 100.0 \% & \qty{36794.5}{\second} & 100.0 \% \\
\informed     & \qty{2537.5}{\second} &  98.7 \% & \qty{2264.1}{\second} &  99.5 \% & \qty{36896.7}{\second} & 100.3 \% \\
\reinformed   & \qty{3096.1}{\second} & 120.5 \% & \qty{2782.2}{\second} & 122.3 \% & \qty{43298.0}{\second} & 117.7 \% \\
\scaffolder   & \qty{4686.4}{\second} & 182.3 \% & \qty{4275.2}{\second} & 187.9 \% & \qty{79689.1}{\second} & 216.6 \% \\
\bottomrule
\end{tabular}
\end{table}

\section{Conclusion} \label{sec:conclusion}

In this work, we improved asymmetric model-based RL under partial observability by studying the effect of asymmetric representation learning on the representation of the observation and on the representation of the additional information.
First, we showed that the \informed uses a loose lower bound on the log-likelihood of the privileged information, and argued that it hinders the learned representation of the additional information in practice.
Second, we proposed the \reinformed that provides a lower bound on the log-likelihood of the privileged information that has a smaller tightness gap.
This tighter lower bound is obtained through a privileged encoder, which is used to train the unprivileged encoder through a latent guidance mechanism.
Third, we demonstrated empirically that this provides an improved learning speed and final performance on most considered partially observable environments compared to the \uninformed and \informed, and that it matches the performance of the \scaffolder with a single world model.

\section*{Acknowledgments} \label{sec:acknowledgments}

Gaspard Lambrechts is a \emph{Marie Skłodowska-Curie Actions} (MSCA) postdoctoral fellow at McGill University in Canada, funded by \emph{Horizon Europe} under grant 101284668.
Gaspard Lambrechts carried out part of this work as a \emph{Fund for Scientific Research} (FNRS) postdoctoral fellow at the University of Liège in Belgium, funded by the \emph{Wallonia-Brussels Federation}.
Gaspard Lambrechts also gratefully acknowledges the financial support of \emph{Wallonia-Brussels International} for his \emph{World Excellence Fellowship}.
Adrien Bolland is a \emph{Fund for Scientific Research} (FNRS) postdoctoral fellow at the University of Liège in Belgium, funded by the \emph{Wallonia-Brussels Federation}.
Daniel Ebi acknowledges financial support by the \emph{German Research Foundation} (DFG) as part of the Research Training Group GRK 2153.
Computational resources have been provided by the \emph{Consortium des Équipements de Calcul Intensif} (CÉCI), funded by the \emph{Fund for Scientific Research} (FNRS) under grant 2502011 and by the \emph{Walloon Region}, including the Tier-1 supercomputer of the \emph{Wallonia-Brussels Federation}, infrastructure funded by the \emph{Walloon Region} under grant~1117545.

\bibliography{references.bib}
\bibliographystyle{rlj}

\newpage \appendix

\section{Extended Related Work} \label{app:related}

In this section, we give an overview of RL under partial observability and asymmetric observability, present the theory of asymmetric RL, and discuss the related literature in multiagent RL.

\subsection{Reinforcement learning under partial observability}

Many RL approaches for partially observable environments have relied on learning a history-dependent policy, usually implemented using a recurrent neural network.
These policies are typically trained with gradient-based RL algorithms using backpropagation through time \citep{bakker2001reinforcement, wierstra2007solving, hausknecht2015deep, heess2015memory, zhang2016learning, zhu2017improving}.
In this case, the hidden state of the recurrent neural network is a statistic of the history, which should ideally summarize all relevant information from the history to act optimally in the future \citep{lambrechts2022recurrent}.
Such a statistic of the history is called a sufficient statistic for optimal control \citep{striebel1965sufficient} or an information state \citep{bertsekas2012dynamic}.
Although these approaches theoretically allow implicit learning of a sufficient statistic, sufficient statistics can also be learned explicitly.
Notably, many works \citep{igl2018deep, buesing2018learning, guo2018neural, gregor2019shaping, han2019variational, guo2020bootstrap, lee2020stochastic, hafner2019learning, hafner2020dream} focused on learning a recurrent statistic that encodes the reward and next observation distribution given the action, a property known as predictive sufficiency \citep{bernardo2009bayesian}.
A recurrent and predictive statistic is indeed proven to be sufficient for optimal control \citep{subramanian2022approximate}.
In practice, the sufficiency objective is usually optimized jointly with the RL objective.

\subsection{Asymmetric reinforcement learning}

As far as asymmetric RL is concerned, we identify four categories of algorithms: imitation learning, asymmetric critics, asymmetric world models, and asymmetric representation learning.
Early approaches proposed to imitate a privileged policy conditioned on the state \citep{choudhury2018data}, or to use an asymmetric critic conditioned on the state \citep{pinto2017asymmetric}.
These heuristic methods initially lacked a theoretical framework, and a recent line of work has focused on proposing theoretically grounded asymmetric learning objectives.
First, imitation learning of a privileged policy was known to be suboptimal \citep{choudhury2018data}, and it was addressed by constraining the privileged policy so that its imitation results in an optimal policy for the partially observable environment \citep{warrington2021robust}.
Similarly, asymmetric actor-critic approaches were proven to provide biased gradients, and an unbiased actor-critic approach was proposed by introducing the history-state value function \citep{baisero2022unbiased}.
Many recent empirical successes of RL have relied on an asymmetric critic \citep{degrave2022magnetic, kaufmann2023champion, vasco2024super, hu2024privileged, hu2025real, durr2026outplaying}.
In model-based RL, several works proposed world model objectives that are proven to provide sufficient statistics of the history for optimal control, by leveraging the state \citep{avalos2024wasserstein} or arbitrary state information \citep{lambrechts2024informed}.
Finally, asymmetric representation learning approaches were proposed to learn sufficient statistics of the history for optimal control, from state samples \citep{wang2023learning, sinha2023asymmetric}.

\subsection{Theory of asymmetric learning}

While a line of work has focused on proposing theoretically grounded asymmetric learning objectives \citep{warrington2021robust, baisero2022unbiased, lambrechts2024informed, wang2023learning}, in the sense that optimizing for these objectives provides optimal policies, a theoretical justification for why these asymmetric objectives help was still missing.
A recent line of work has focused on finding a theoretical justification for asymmetric actor-critic algorithms.
A theoretical justification was proposed by adapting a finite-time convergence analysis for actor-critic algorithms with fixed agent state and linear function approximator \citep{cayci2024finite} to the asymmetric setting \citep{lambrechts2025theoretical}.
The comparison of these finite-time bounds highlighted an advantage of asymmetric learning in the presence of aliasing in the agent state space.
An alternative finite-time bound was provided for a tabular belief-weighted asymmetric actor-critic, highlighting again the gains from the privileged information \citep{cai2024provable}.
However, both these theoretical works assumed privileged access to the full state during training.
Recent work proposed to extend the asymmetric actor-critic algorithm to arbitrary additional state information \citep{ebi2025leveraging}.
Then, by adapting a finite-time convergence proof of recurrent actor-critic \citep{cayci2024recurrent}, several hypotheses for the benefits of asymmetric recurrent actor-critic are discussed \citep{ebi2025theoretical}.
As far as asymmetric world models and asymmetric representation learning are concerned, a detailed theoretical justification for their effectiveness is still missing.

\subsection{Centralized training for decentralized execution}

In multiagent RL, exploiting additional information available during training was extensively studied under the centralized training for decentralized execution (CTDE) framework \citep{oliehoek2008optimal, amato2024introduction}.
It is also worth noting that CTDE was definitely an inspiration for many asymmetric RL methods for partial observability.
In CTDE, it is assumed that the histories of all agents, or even the environment state, are available to all agents during training.
To exploit this additional information, several asymmetric actor-critic approaches have been developed by leveraging an asymmetric critic conditioned on all histories, including COMA \citep{foerster2018counterfactual}, MADDPG \citep{lowe2017multi}, M3DDPG \citep{li2019robust} and R-MADDPG \citep{wang2020partially}.
While efficient in practice, \citet{lyu2022deeper} showed that these asymmetric actor-critic approaches provide biased gradient estimates, which generalizes results developed in the asymmetric learning literature \citep{baisero2022unbiased} to the multi-agent setting.
In the cooperative CTDE setting, another line of work focuses on value decomposition to learn a utility function for each agent, including QMIX \citep{rashid2018qmix}, QVMix \citep{leroy2021qvmix} and QPLEX \citep{wang2021qplex}.
These approaches use the additional information to modulate the contribution of each utility function in the global value function, while ensuring that maximizing the local utility functions also maximize the global value function, a property known as individual global maximum (IGM).
Other methods relax this IGM requirement but still condition the value function on all histories, including QTRAN \citep{son2019qtran} and WQMix \citep{rashid2020weighted}.
\citet{hong2022rethinking} established that the IGM decomposition is not attainable in the general case.

\section{Informed Evidence Lower Bound}

In \autoref{app:informed_elbo}, we introduce the informed ELBO as a lower bound on the log-likelihood.
In \autoref{app:informed_gap}, we show that this lower bound is loose in general.

\subsection{Informed Evidence Lower Bound} \label{app:informed_elbo}

First, let us recall the prior distribution and the variational distribution of the informed RSSM:
\begin{align}
    q_\phi(e_{0:t} | h_t) &= \textstyle \prod_{k=0}^{t} q_\phi^e(e_k | z_k, o_k), \\
    p_\phi(e_{0:t}, \hat e_{t+1} | h_t, a_t) &= q_\phi(e_{0:t} | h_t) p_\phi^{\hat{e}}(\hat e_{t+1} | z_{t+1}), \\
    q_\phi(e_{0:t}, e_{t+1} | h_t, a_t, o_{t+1}) &= q_\phi(e_{0:t} | h_t) q_\phi^e(e_{t+1} | z_{t+1}, o_{t+1})
    = q_\phi(e_{0:t+1} | h_{t+1}),
\end{align}
where $z_{k+1} = u_\phi(z_{k}, e_{k}, a_{k})$ for $k \leq t$.
Starting from the log-likelihood $\log p_\phi(r_t, i_{t+1} | h_t, a_t)$ of the LVM at equation~\eqref{eq:lvm}, we have:
\begin{align}
    &\log p_\phi(r_t, i_{t+1} | h_t, a_t)
    \nonumber \\
    &\quad
    = \log \mathop{\mathbb{E}}_{p_\phi(e_{0:t}, \hat{e}_{t+1} | h_t, a_t)} \left[
            p_\phi(r_t, i_{t+1} | z_{t+1}, \hat{e}_{t+1})
    \right]
    \\
    &\quad
    = \log
        \mathop{\mathbb{E}}_{\wt{O}(o_{t+1} | i_{t+1})} \left[
            \mathop{\mathbb{E}}_{p_\phi(e_{0:t}, \hat{e}_{t+1} | h_t, a_t)} \left[
                \frac{
                    p_\phi(r_t, i_{t+1} | z_{t+1}, \hat{e}_{t+1})
                    q_\phi^e(\hat{e}_{t+1} | z_{t+1}, o_{t+1})
                }{
                    q_\phi^e(\hat{e}_{t+1} | z_{t+1}, o_{t+1})
                }
            \right]
    \right]
    \\
    &\quad
    = \log \mathop{\mathbb{E}}_{\wt{O}(o_{t+1} | i_{t+1})} \left[
        \mathop{\mathbb{E}}_{p_\phi(e_{0:t}, {e}_{t+1} | h_t, a_t)} \left[
                \frac{
                    p_\phi(r_t, i_{t+1} | z_{t+1}, {e}_{t+1})
                    q_\phi^e({e}_{t+1} | z_{t+1}, o_{t+1})
                }{
                    q_\phi^e({e}_{t+1} | z_{t+1}, o_{t+1})
                }
        \right]
    \right]
    \\
    &\quad
    = \log \mathop{\mathbb{E}}_{\wt{O}(o_{t+1} | i_{t+1})} \left[
        \mathop{\mathbb{E}}_{p_\phi(e_{0:t}, {e}_{t+1} | h_t, a_t)} \left[
                \frac{
                    p_\phi(r_t, i_{t+1} | z_{t+1}, {e}_{t+1})
                    q_\phi(e_{0:t} | h_t)
                    q_\phi^e({e}_{t+1} | z_{t+1}, o_{t+1})
                }{
                    q_\phi(e_{0:t} | h_t)
                    q_\phi^e({e}_{t+1} | z_{t+1}, o_{t+1})
                }
        \right]
    \right]
    \\
    &\quad
    = \log \mathop{\mathbb{E}}_{\wt{O}(o_{t+1} | i_{t+1})} \left[
        \mathop{\mathbb{E}}_{p_\phi(e_{0:t}, {e}_{t+1} | h_t, a_t)} \left[
                \frac{
                    p_\phi(r_t, i_{t+1} | z_{t+1}, {e}_{t+1})
                    q_\phi(e_{0:t}, e_{t+1} | h_t, a_t, o_{t+1})
                }{
                    q_\phi(e_{0:t} | h_t)
                    q_\phi^e({e}_{t+1} | z_{t+1}, o_{t+1})
                }
        \right]
    \right]
    \\
    &\quad
    = \log \mathop{\mathbb{E}}_{\wt{O}(o_{t+1} | i_{t+1})} \left[
        \mathop{\mathbb{E}}_{q_\phi(e_{0:t}, e_{t+1} | h_t, a_t, o_{t+1})} \left[
                \frac{
                    p_\phi(r_t, i_{t+1} | z_{t+1}, {e}_{t+1})
                    p_\phi(e_{0:t}, {e}_{t+1} | h_t, a_t)
                }{
                    q_\phi(e_{0:t} | h_t)
                    q_\phi^e({e}_{t+1} | z_{t+1}, o_{t+1})
                }
        \right]
    \right]
    \\
    &\quad
    = \log \mathop{\mathbb{E}}_{\wt{O}(o_{t+1} | i_{t+1})} \left[
        \mathop{\mathbb{E}}_{q_\phi(e_{0:t}, e_{t+1} | h_t, a_t, o_{t+1})} \left[
                \frac{
                    p_\phi(r_t, i_{t+1} | z_{t+1}, {e}_{t+1})
                    q_\phi(e_{0:t} | h_t) p_\phi^{\hat{e}}({e}_{t+1} | z_{t+1})
                }{
                    q_\phi(e_{0:t} | h_t)
                    q_\phi^e({e}_{t+1} | z_{t+1}, o_{t+1})
                }
        \right]
    \right]
    \\
    &\quad
    = \log \mathop{\mathbb{E}}_{\wt{O}(o_{t+1} | i_{t+1})} \left[
        \mathop{\mathbb{E}}_{q_\phi(e_{0:t}, e_{t+1} | h_t, a_t, o_{t+1})} \left[
                \frac{
                    p_\phi(r_t, i_{t+1} | z_{t+1}, e_{t+1})
                    p_\phi^{\hat{e}}(e_{t+1} | z_{t+1})
                }{
                    q_\phi^e(e_{t+1} | z_{t+1}, o_{t+1})
                }
        \right]
    \right]
    \\
    &\quad
    \geq \mathop{\mathbb{E}}_{\wt{O}(o_{t+1} | i_{t+1})} \left[
        \mathop{\mathbb{E}}_{q_\phi(e_{0:t}, e_{t+1} | h_t, a_t, o_{t+1})} \left[
                \log \frac{
                    p_\phi(r_t, i_{t+1} | z_{t+1}, e_{t+1})
                    p_\phi^{\hat{e}}(e_{t+1} | z_{t+1})
                }{
                    q_\phi^e(e_{t+1} | z_{t+1}, o_{t+1})
                }
        \right]
    \right] \label{eq:informed_elbo_frac}
    \\
    &\quad
    = \mathop{\mathbb{E}}_{\wt{O}(o_{t+1} | i_{t+1})} \Bigg[
        \mathop{\mathbb{E}}_{q_\phi(e_{0:t}, e_{t+1} | h_t, a_t, o_{t+1})} \Bigg[
                \log p_\phi^r(r_t | z_{t+1}, e_{t+1})
                +
                \log p_\phi^i(i_{t+1} | z_{t+1}, e_{t+1})
            \nonumber \\
            &\qquad\qquad\qquad\qquad\qquad\qquad\qquad
            - \operatorname{KL}_{e_{t+1}'}(
                q_\phi^e(e_{t+1}' | z_{t+1}, o_{t+1})
                \parallel
                p_\phi^{\hat{e}}(e_{t+1}' | z_{t+1})
            )
        \Bigg]
    \Bigg]
    \\
    &\quad
    = \wt{\operatorname{ELBO}}_\phi(r_t, i_{t+1} | h_t, a_t).
\end{align}

\subsection{Informed Tightness Gap} \label{app:informed_gap}

Starting from equation~\eqref{eq:informed_elbo_frac}, we can develop the ELBO further.
We have:
\begin{align}
    &\wt{\operatorname{ELBO}}_\phi(r_t, i_{t+1} | h_t, a_t)
    \nonumber \\
    &\quad
    = \mathop{\mathbb{E}}_{\wt{O}(o_{t+1} | i_{t+1})} \Bigg[
        \mathop{\mathbb{E}}_{q_\phi(e_{0:t}, e_{t+1} | h_t, a_t, o_{t+1})} \Bigg[
                \log
                \frac{
                    p_\phi(r_t, i_{t+1} | z_{t+1}, e_{t+1})
                    p_\phi^{\hat{e}}(e_{t+1} | z_{t+1})
                }{
                    q_\phi^e(e_{t+1} | z_{t+1}, o_{t+1})
                }
            \Bigg]
    \Bigg]
    \\
    &\quad
    = \mathop{\mathbb{E}}_{\wt{O}(o_{t+1} | i_{t+1})} \Bigg[
        \mathop{\mathbb{E}}_{q_\phi(e_{0:t+1} | h_t, a_t, o_{t+1})} \Bigg[
                \log
                \frac{
                    p_\phi(r_t, i_{t+1} | z_{t+1}, e_{t+1})
                    p_\phi^{\hat{e}}(e_{t+1} | z_{t+1})
                }{
                    q_\phi^e(e_{t+1} | z_{t+1}, o_{t+1})
                }
            \Bigg]
    \Bigg]
    \\
    &\quad
    = \mathop{\mathbb{E}}_{\wt{O}(o_{t+1} | i_{t+1})} \Bigg[
        \mathop{\mathbb{E}}_{q_\phi(e_{0:t+1} | h_t, a_t, o_{t+1})} \Bigg[
                \log
                \frac{
                    p_\phi(r_t, i_{t+1} | z_{t+1}, e_{t+1})
                    q_\phi(e_{0:t} | h_t) p_\phi^{\hat{e}}(e_{t+1} | z_{t+1})
                }{
                    q_\phi(e_{0:t} | h_t) q_\phi^e(e_{t+1} | z_{t+1}, o_{t+1})
                }
        \Bigg]
    \Bigg]
    \\
    &\quad
    = \mathop{\mathbb{E}}_{\wt{O}(o_{t+1} | i_{t+1})} \Bigg[
        \mathop{\mathbb{E}}_{q_\phi(e_{0:t+1} | h_t, a_t, o_{t+1})} \Bigg[
                \log \frac{
                    p_\phi(r_t, i_{t+1} | z_{t+1}, e_{t+1})
                    p_\phi(e_{0:t+1} | h_t, a_t)
                }{
                    q_\phi(e_{0:t+1} | h_t, a_t, o_{t+1})
                }
        \Bigg]
    \Bigg]
    \\
    &\quad
    = \mathop{\mathbb{E}}_{\wt{O}(o_{t+1} | i_{t+1})} \Bigg[
        \mathop{\mathbb{E}}_{q_\phi(e_{0:t+1} | h_t, a_t, o_{t+1})} \Bigg[
                \log \frac{
                    p_\phi(e_{0:t+1}, r_t, i_{t+1} | h_t, a_t)
                }{
                    q_\phi(e_{0:t+1} | h_t, a_t, o_{t+1})
                }
        \Bigg]
    \Bigg]
    \\
    &\quad
    = \mathop{\mathbb{E}}_{\wt{O}(o_{t+1} | i_{t+1})} \Bigg[
        \mathop{\mathbb{E}}_{q_\phi(e_{0:t+1} | h_t, a_t, o_{t+1})} \Bigg[
                \log \frac{
                    p_\phi(r_t, i_{t+1} | h_t, a_t)
                    p_\phi(e_{0:t+1} | h_t, a_t, r_t, i_{t+1})
                }{
                    q_\phi(e_{0:t+1} | h_t, a_t, o_{t+1})
                }
        \Bigg]
    \Bigg]
    \\
    &\quad
    =
    \log p_\phi(r_t, i_{t+1} | h_t, a_t)
    \nonumber \\
    &\qquad\qquad
    -
    \mathop{\mathbb{E}}_{\wt{O}(o_{t+1} | i_{t+1})} \big[
        \operatorname{KL}_{e_{0:t+1}}(q_\phi(e_{0:t+1} | h_t, a_t, o_{t+1}) \parallel p_\phi(e_{0:t+1} | h_t, a_t, r_t, i_{t+1}))
    \big],
\end{align}
where $p_\phi(e_{0:t+1} | h_t, a_t, r_t, i_{t+1})$ is the true posterior of the LVM.

\section{Reinformed Evidence Lower Bound}

In \autoref{app:reinformed_elbo}, we introduce the reinformed ELBO as a lower bound on the log-likelihood.
In \autoref{app:reinformed_gap}, we characterize the tightness gap of this lower bound.

\subsection{Reinformed Evidence Lower Bound} \label{app:reinformed_elbo}

First, let us recall the prior distribution and the variational distribution of the reinformed RSSM:
\begin{align}
    q_\phi(e_{0:t} | h_t) &= \textstyle \prod_{k=0}^{t} q_\phi^e(e_k | z_k, o_k), \\
    p_\phi(e_{0:t}, \hat e_{t+1} | h_t, a_t) &= q_\phi(e_{0:t} | h_t) p_\phi^{\hat{e}}(\hat e_{t+1} | z_{t+1}), \\
    q_\phi(e_{0:t}, \tilde e_{t+1} | h_t, a_t, i_{t+1}) &= q_\phi(e_{0:t} | h_t) q_\phi^{\tilde e}(\tilde e_{t+1} | z_{t+1}, i_{t+1}),
\end{align}
where $z_{k+1} = u_\phi(z_{k}, e_{k}, a_{k})$ for $k \leq t$.
Starting from the log-likelihood $\log p_\phi(r_t, i_{t+1} | h_t, a_t)$ of the LVM at equation~\eqref{eq:lvm}, we have:
\begin{align}
    &\log p_\phi(r_t, i_{t+1} | h_t, a_t)
    \nonumber \\
    &\quad
    = \log \mathop{\mathbb{E}}_{p_\phi(e_{0:t}, \hat{e}_{t+1} | h_t, a_t)} \left[
            p_\phi(r_t, i_{t+1} | z_{t+1}, \hat{e}_{t+1})
    \right]
    \\
    &\quad
    = \log
            \mathop{\mathbb{E}}_{p_\phi(e_{0:t}, \hat{e}_{t+1} | h_t, a_t)} \left[
                \frac{
                    p_\phi(r_t, i_{t+1} | z_{t+1}, \hat{e}_{t+1})
                    q_\phi^{\tilde e}(\hat{e}_{t+1} | z_{t+1}, i_{t+1})
                }{
                    q_\phi^{\tilde e}(\hat{e}_{t+1} | z_{t+1}, i_{t+1})
                }
            \right]
    \\
    &\quad
    = \log
        \mathop{\mathbb{E}}_{p_\phi(e_{0:t}, \tilde{e}_{t+1} | h_t, a_t)} \left[
                \frac{
                    p_\phi(r_t, i_{t+1} | z_{t+1}, \tilde{e}_{t+1})
                    q_\phi^{\tilde e}(\tilde{e}_{t+1} | z_{t+1}, i_{t+1})
                }{
                    q_\phi^{\tilde e}(\tilde{e}_{t+1} | z_{t+1}, i_{t+1})
                }
        \right]
    \\
    &\quad
    = \log
        \mathop{\mathbb{E}}_{p_\phi(e_{0:t}, \tilde{e}_{t+1} | h_t, a_t)} \left[
                \frac{
                    p_\phi(r_t, i_{t+1} | z_{t+1}, \tilde{e}_{t+1})
                    q_\phi(e_{0:t} | h_t)
                    q_\phi^{\tilde e}(\tilde{e}_{t+1} | z_{t+1}, i_{t+1})
                }{
                    q_\phi(e_{0:t} | h_t)
                    q_\phi^{\tilde e}(\tilde{e}_{t+1} | z_{t+1}, i_{t+1})
                }
        \right]
    \\
    &\quad
    = \log
        \mathop{\mathbb{E}}_{p_\phi(e_{0:t}, \tilde{e}_{t+1} | h_t, a_t)} \left[
                \frac{
                    p_\phi(r_t, i_{t+1} | z_{t+1}, \tilde{e}_{t+1})
                    q_\phi(e_{0:t}, \tilde e_{t+1} | h_t, a_t, i_{t+1})
                }{
                    q_\phi(e_{0:t} | h_t)
                    q_\phi^{\tilde e}(\tilde{e}_{t+1} | z_{t+1}, i_{t+1})
                }
        \right]
    \\
    &\quad
    = \log
        \mathop{\mathbb{E}}_{q_\phi(e_{0:t}, \tilde e_{t+1} | h_t, a_t, i_{t+1})} \left[
                \frac{
                    p_\phi(r_t, i_{t+1} | z_{t+1}, \tilde{e}_{t+1})
                    p_\phi(e_{0:t}, \tilde{e}_{t+1} | h_t, a_t)
                }{
                    q_\phi(e_{0:t} | h_t)
                    q_\phi^{\tilde e}(\tilde{e}_{t+1} | z_{t+1}, i_{t+1})
                }
        \right]
    \\
    &\quad
    = \log
        \mathop{\mathbb{E}}_{q_\phi(e_{0:t}, \tilde e_{t+1} | h_t, a_t, i_{t+1})} \left[
                \frac{
                    p_\phi(r_t, i_{t+1} | z_{t+1}, \tilde{e}_{t+1})
                    q_\phi(e_{0:t} | h_t) p_\phi^{\hat{e}}(\tilde{e}_{t+1} | z_{t+1})
                }{
                    q_\phi(e_{0:t} | h_t)
                    q_\phi^{\tilde e}(\tilde{e}_{t+1} | z_{t+1}, i_{t+1})
                }
        \right]
    \\
    &\quad
    = \log
        \mathop{\mathbb{E}}_{q_\phi(e_{0:t}, \tilde{e}_{t+1} | h_t, a_t, i_{t+1})} \left[
                \frac{
                    p_\phi(r_t, i_{t+1} | z_{t+1}, \tilde e_{t+1})
                    p_\phi^{\hat{e}}(\tilde e_{t+1} | z_{t+1})
                }{
                    q_\phi^{\tilde e}(\tilde e_{t+1} | z_{t+1}, i_{t+1})
                }
        \right]
    \\
    &\quad
    \geq
        \mathop{\mathbb{E}}_{q_\phi(e_{0:t}, \tilde e_{t+1} | h_t, a_t, i_{t+1})} \left[
                \log \frac{
                    p_\phi(r_t, i_{t+1} | z_{t+1}, \tilde e_{t+1})
                    p_\phi^{\hat{e}}(\tilde e_{t+1} | z_{t+1})
                }{
                    q_\phi^{\tilde e}(\tilde e_{t+1} | z_{t+1}, i_{t+1})
                }
        \right] \label{eq:reinformed_elbo_frac}
    \\
    &\quad
    =
        \mathop{\mathbb{E}}_{q_\phi(e_{0:t}, \tilde e_{t+1} | h_t, a_t, i_{t+1})} \Bigg[
                \log p_\phi^r(r_t | z_{t+1}, \tilde e_{t+1})
                +
                \log p_\phi^i(i_{t+1} | z_{t+1}, \tilde e_{t+1})
            \nonumber \\
            &\qquad\qquad\qquad\qquad\qquad\qquad\qquad
            - \operatorname{KL}_{\tilde e_{t+1}'}(
                q_\phi^{\tilde e}(\tilde e_{t+1}' | z_{t+1}, i_{t+1})
                \parallel
                p_\phi^{\hat{e}}(\tilde e_{t+1}' | z_{t+1})
            )
        \Bigg]
    \\
    &\quad
    = \operatorname{ELBO}_\phi(r_t, i_{t+1} | h_t, a_t).
\end{align}

\subsection{Reinformed Tightness Gap} \label{app:reinformed_gap}

Starting from equation~\eqref{eq:reinformed_elbo_frac}, we can develop the ELBO further.
We have:
\begin{align}
    &
    \operatorname{ELBO}_\phi(r_t, i_{t+1} | h_t, a_t)
    \nonumber \\
    &\quad
    =
        \mathop{\mathbb{E}}_{q_\phi(e_{0:t}, \tilde e_{t+1} | h_t, a_t, i_{t+1})} \Bigg[
                \log
                \frac{
                    p_\phi(r_t, i_{t+1} | z_{t+1}, \tilde e_{t+1})
                    p_\phi^{\hat{e}}(\tilde e_{t+1} | z_{t+1})
                }{
                    q_\phi^{\tilde e}(\tilde e_{t+1} | z_{t+1}, i_{t+1})
                }
        \Bigg]
    \\
    &\quad
    =
        \mathop{\mathbb{E}}_{q_\phi(e_{0:t}, e_{t+1} | h_t, a_t, i_{t+1})} \Bigg[
                \log
                \frac{
                    p_\phi(r_t, i_{t+1} | z_{t+1}, e_{t+1})
                    p_\phi^{\hat{e}}(e_{t+1} | z_{t+1})
                }{
                    q_\phi^{\tilde e}(e_{t+1} | z_{t+1}, i_{t+1})
                }
        \Bigg]
    \\
    &\quad
    =
        \mathop{\mathbb{E}}_{q_\phi(e_{0:t+1} | h_t, a_t, i_{t+1})} \Bigg[
                \log
                \frac{
                    p_\phi(r_t, i_{t+1} | z_{t+1}, e_{t+1})
                    p_\phi^{\hat{e}}(e_{t+1} | z_{t+1})
                }{
                    q_\phi^{\tilde e}(e_{t+1} | z_{t+1}, i_{t+1})
                }
        \Bigg]
    \\
    &\quad
    =
        \mathop{\mathbb{E}}_{q_\phi(e_{0:t+1} | h_t, a_t, i_{t+1})} \Bigg[
                \log
                \frac{
                    p_\phi(r_t, i_{t+1} | z_{t+1}, e_{t+1})
                    q_\phi(e_{0:t} | h_t) p_\phi^{\hat{e}}(e_{t+1} | z_{t+1})
                }{
                    q_\phi(e_{0:t} | h_t) q_\phi^{\tilde e}(e_{t+1} | z_{t+1}, i_{t+1})
                }
        \Bigg]
    \\
    &\quad
    =
        \mathop{\mathbb{E}}_{q_\phi(e_{0:t+1} | h_t, a_t, i_{t+1})} \Bigg[
                \log
                \frac{
                    p_\phi(r_t, i_{t+1} | z_{t+1}, e_{t+1})
                    p_\phi(e_{0:t+1} | h_t, a_t)
                }{
                    q_\phi(e_{0:t+1} | h_t, a_t, i_{t+1})
                }
        \Bigg]
    \\
    &\quad
    =
        \mathop{\mathbb{E}}_{q_\phi(e_{0:t+1} | h_t, a_t, i_{t+1})} \Bigg[
                \log \frac{
                    p_\phi(e_{0:t+1}, r_t, i_{t+1} | h_t, a_t)
                }{
                    q_\phi(e_{0:t+1} | h_t, a_t, i_{t+1})
                }
        \Bigg]
    \\
    &\quad
    =
        \mathop{\mathbb{E}}_{q_\phi(e_{0:t+1} | h_t, a_t, i_{t+1})} \Bigg[
                \log \frac{
                    p_\phi(r_t, i_{t+1} | h_t, a_t)
                    p_\phi(e_{0:t+1} | h_t, a_t, r_t, i_{t+1})
                }{
                    q_\phi(e_{0:t+1} | h_t, a_t, i_{t+1})
                }
        \Bigg]
    \\
    &\quad
    =
    \log p_\phi(r_t, i_{t+1} | h_t, a_t)
    \nonumber \\
    &\qquad\qquad
    -
        \operatorname{KL}_{e_{0:t+1}}(q_\phi(e_{0:t+1} | h_t, a_t, i_{t+1}) \parallel p_\phi(e_{0:t+1} | h_t, a_t, r_t, i_{t+1})),
\end{align}
where $p_\phi(e_{0:t+1} | h_t, a_t, r_t, i_{t+1})$ is the true posterior of the LVM.

\section{Experimental Details} \label{app:details}

For the \uninformed \citep{hafner2025mastering}, \informed \citep{lambrechts2024informed} and \scaffolder \citep{hu2024privileged} algorithms, we use the original implementations and hyperparameters.
Similarly, the \reinformed algorithm is based on the original \informed implementation and hyperparameters.

As far as the training objectives of the unprivileged encoder and privileged encoder are concerned, we use the same hyperparameters as the observational encoder in the \uninformed and \informed.
More precisely, we use KL balancing with $\beta_\text{dyn} = 0.5$ and $\beta_\text{rep} = 0.1$ and free bits of $1$ nat for both encoders.
By denoting the stop-gradient operator with $\operatorname{sg}(\cdot)$, we have:
\begin{align}
    \widehat{\operatorname{KL}}_{e_t}(q^{\tilde e}_\phi(e_t | z_t, i_t) \parallel p_\phi^{\hat e}(e_t | z_t))
    &=
    \beta_\text{dyn} \operatorname{max}(1, \operatorname{KL}_{e_t}(\operatorname{sg}(q^{\tilde e}_\phi(e_t | z_t, i_t)) \parallel p_\phi^{\hat e}(e_t | z_t)))
    \nonumber \\
    &\qquad\quad+
    \beta_\text{rep} \operatorname{max}(1, \operatorname{KL}_{e_t}(q^{\tilde e}_\phi(e_t | z_t, i_t) \parallel \operatorname{sg}(p_\phi^{\hat e}(e_t | z_t))))
    \\
    \widehat{\operatorname{KL}}_{e_t}(q^{e}_\phi(e_t | z_t, o_t) \parallel p_\phi^{\hat e}(e_t | z_t))
    &=
    \beta_\text{dyn} \operatorname{max}(1, \operatorname{KL}_{e_t}(\operatorname{sg}(q^{e}_\phi(e_t | z_t, o_t)) \parallel p_\phi^{\hat e}(e_t | z_t)))
    \nonumber \\
    &\qquad\quad+
    \beta_\text{rep} \operatorname{max}(1, \operatorname{KL}_{e_t}(q^{e}_\phi(e_t | z_t, o_t) \parallel \operatorname{sg}(p_\phi^{\hat e}(e_t | z_t)))).
\end{align}

In all environments, we use $i_t = (o_t, o_t^+)$ for the Reinformed Dreamer even if the additional information $o_t^+$ is known to be a Markovian state, as we found it slightly more effective.

\end{document}